\documentclass[journal]{IEEEtran}      

\IEEEoverridecommandlockouts           

\usepackage{graphicx}
\usepackage{graphics}
\usepackage{amsmath,bm}
\usepackage{amssymb}
\usepackage{soul}
\usepackage{algorithm}
\usepackage{algpseudocode}
\usepackage{cite}
\usepackage{hyperref}
\usepackage[caption=false]{subfig}
\usepackage{booktabs}
\usepackage{siunitx}
\usepackage{float}
\usepackage{multirow}
\usepackage{bm}
\usepackage{color}
\usepackage{changes}
\usepackage{balance}
\hypersetup{hidelinks}
\usepackage{url}

\algdef{SE}[DOWHILE]{Do}{doWhile}{\algorithmicdo}[1]{\algorithmicwhile\ #1}%

\newcommand{\mathactivatecomma}{%
  \begingroup\lccode`~=`\,
  \lowercase{\endgroup\edef~}{\mathchar\the\mathcode`\,\penalty0 }}

\algnewcommand{\Initialize}[1]{%
  \State \textbf{Initialize: $i \in \mathcal{V}$}
  \Statex \hspace*{\algorithmicindent}\parbox[t]{.8\linewidth}{\raggedright #1}
}

\algnewcommand{\Iteration}[1]{%
  \State \textbf{Iteration $(k\geq 0)$: $i \in \mathcal{V}$}
  \Statex \hspace*{\algorithmicindent}\parbox[t]{.8\linewidth}{\raggedright #1}
}

\algnewcommand{\Output}[1]{
  \State \textbf{Output: $i \in \mathcal{V}$}
  \Statex \hspace*{\algorithmicindent}\parbox[t]{.8\linewidth}{\raggedright #1}
}

\title{\LARGE \bf
LRT: An Efficient Low-Light Restoration Transformer for Dark Light Field Images}

\author{Shansi Zhang, Nan Meng, and Edmund Y. Lam

\thanks{S. Zhang and E.Y. Lam are with the Department of Electrical and Electronic Engineering, The University of Hong Kong, Pokfulam, Hong Kong SAR, China {\tt\small e-mail: sszhang@eee.hku.hk, elam@eee.hku.hk}.}
\thanks{N. Meng is with the Li Ka Shing Faculty of Medicine, The University of Hong Kong, Pokfulam, Hong Kong SAR, China.}
}

\usepackage{color}
\definecolor{darkspringgreen}{rgb}{0.09, 0.45, 0.27}

\begin{document}

\maketitle
\thispagestyle{empty}
\pagestyle{empty}

\begin{abstract}
Light field (LF) images containing information for multiple views have numerous applications, which can be severely affected by low-light imaging. Recent learning-based methods for low-light enhancement have some disadvantages, such as a lack of noise suppression, complex training process and poor performance in extremely low-light conditions. To tackle these deficiencies while fully utilizing the multi-view information, we propose an efficient Low-light Restoration Transformer (LRT) for LF images, with multiple heads to perform intermediate tasks within a single network, including denoising, luminance adjustment, refinement and detail enhancement, achieving progressive restoration from small scale to full scale. Moreover, we design an angular transformer block with an efficient view-token scheme to model the global angular dependencies, and a multi-scale spatial transformer block to encode the multi-scale local and global information within each view. To address the issue of insufficient training data, we formulate a synthesis pipeline by simulating the major noise sources with the estimated noise parameters of LF camera. Experimental results demonstrate that our method achieves the state-of-the-art performance on low-light LF restoration with high efficiency. 
\end{abstract}


\section{Introduction}\label{sec:introduction}
Light field (LF) cameras can capture both the intensities and directions of the light rays, which enables multi-view imaging to obtain an array of sub-aperture images (SAIs) and brings many applications, such as post-capture refocusing~\cite{Fiss2014,Wang2019a}, depth estimation~\cite{Shi2019,Jin2022}, de-occlusion~\cite{Wang2020a,Li2021} and saliency detection~\cite{Zhang2020a,Zhang2020b}. However, these applications are susceptible to the degraded LF images caused by low-light imaging, which leads to missing contents and serious noise. Simply increasing the ISO and exposure time are not helpful, since they also increase the noise level and introduce blurriness. Therefore, low-light enhancement algorithms are needed to recover visibility and suppress noise for better LF applications.

Nowadays, deep learning-based methods have shown improved performance and efficiency in low-light enhancement for both the conventional images and LF images. Some methods~\cite{Lv2018,Jiang2019,Lamba2021,Lamba2022} design a network to learn a direct mapping from low-light images to normal images. However, it is difficult to achieve denoising, luminance enhancement and color recovery with a direct image-to-image mapping, resulting in poor performance in extremely low-light conditions. Some methods~\cite{Wang2019b,Guo2020,Ma2022} focus on the illumination adjustment using estimated illumination map or light-enhancement curve. However, they do not incorporate denoising, which limits their applications in real low-light scenarios with non-negligible noise. Some methods~\cite{Zhang2019,Zhang2021c,Wu2022,Zhang2021b,Zhang2022} apply the Retinex theory~\cite{Land1977} by training separate networks to achieve decomposition and enhancement. However, the training process is tedious and the inference efficiency is usually lower than an end-to-end network. To address the above deficiencies, we develop an efficient framework with multiple heads to explicitly perform denoising, luminance enhancement, refinement and detail preservation within a single network, achieving low-light restoration progressively from small scale to larger scales. All these tasks can be learned simultaneously through an end-to-end training.

Different from the conventional images, LF images contain information for multiple views with rich geometric cues. The complementary information across different views is critical for better LF restoration. Some methods~\cite{Jin2020,Lamba2021,Zhang2021a} leverage multiple surrounding views to restore the central view. Nevertheless, only one view can be restored at each forward process, resulting in low efficiency. Some methods~\cite{Yeung2019,Zhang2021b,Liang2022} apply convolution or self-attention operations on the macro-pixels to extract angular features. Even if they can restore all the views synchronously, frequent conversions between the SAI-array pattern and macro-pixel pattern are required, which complicates the feed-forward process and also affects the efficiency. To tackle these issues, we propose an efficient angular transformer block to fully incorporate the global angular dependencies and enable synchronous restoration for all the views without any pattern conversion. To effectively extract features within each view, we propose a spatial transformer block with multi-scale self-attention, which requires much less computation than the common global self-attention and enables more effective global and multi-scale local feature encoding compared to the methods in~\cite{Liu2021b,Wang2022b,Wang2022a,Ren2022}.

With the above-mentioned multi-head framework, angular and spatial transformer blocks, we develop an efficient Low-light Restoration Transformer (LRT) for LF images. In addition, paired low-light/normal LF images are required to train the network. However, there are insufficient data available, and it is difficult and costly to collect a large dataset with aligned image pairs. To solve this problem, we formulate a synthesis pipeline by modeling the sensor noise and estimating the noise parameters of LF camera to synthesize more realistic low-light LF images. Our main contributions are summarized as follows:
\begin{itemize}
\item We develop a transformer-based network with multiple heads to perform denoising, luminance adjustment, refinement and detail enhancement progressively, which separates the complex task into several intermediate tasks for better restoration while ensuring high inference efficiency. 
\item We propose an angular transformer block with an efficient view-token scheme to fully utilize the information of all the views for restoring each individual view. We then propose a spatial transformer block with multi-scale local and global self-attention to encode rich spatial information within each view. 
\item Experimental results on the real low-light LF images demonstrate that our LRT outperforms the state-of-the-art low-light enhancement methods with better quantitative and qualitative results.
\end{itemize}

\begin{figure*}[t]
\centering
\includegraphics[width=1\textwidth]{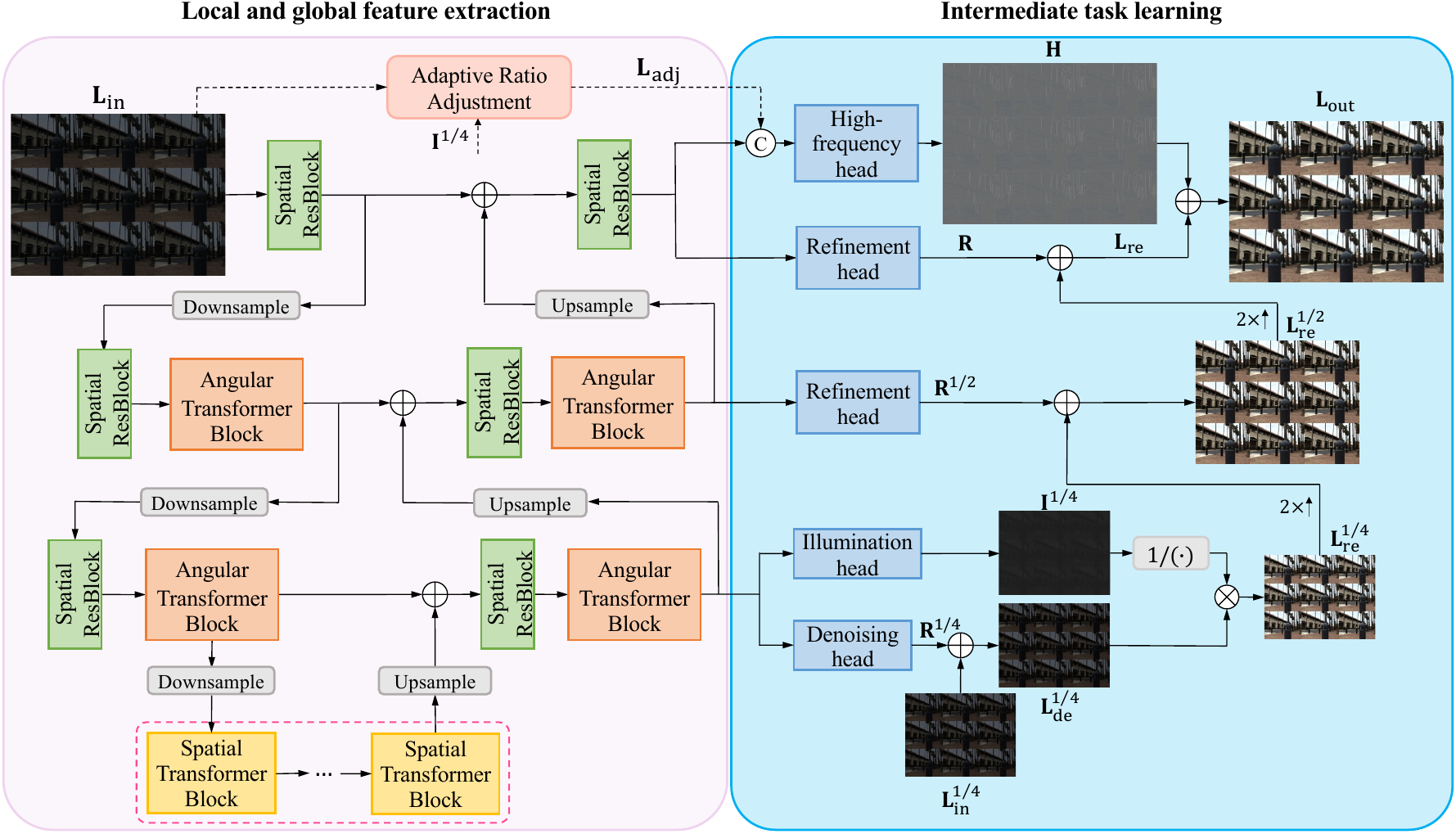} 
\caption{Illustration of our proposed LRT. It leverages spatial ResBlocks and spatial transformer blocks to extract the local and global features within each view, and angular transformer blocks to explore the dependencies among all the views. It contains multiple heads to achieve the intermediate tasks, with denoising and illumination estimation in the $\frac{1}{4}$-scale branch, refinement and detail enhancement in the $\frac{1}{2}$- and full-scale branches.}
\label{fig:architecture}
\end{figure*}

\section{Related work}\label{sec:related work}

\subsection{Low-light Image Enhancement}
Earlier model-based methods~\cite{Fu2016,Guo2017,Li2018,Ren2018} for low-light image enhancement usually applied the Retinex theory to decompose a low-light image into its illumination and reflectance, and then adjust the illumination and suppress noise with traditional approaches. In particular, Lin et al.~\cite{Lin2022} proposed to use edge-preserving filters with the plug-and-play technique to improve illumination, and adopt pixel-wise weights based on variance and image gradients to suppress noise while preserving details in the reflectance layer. These methods usually rely on the carefully designed priors and constraints, and their inference speeds are relatively low. 

With the prevalence of deep learning, more and more methods improved the performance and efficiency by training a deep convolutional neural network (CNN). Lv et al.~\cite{Lv2018} proposed a multi-branch network that fuses the output of multiple subnets to obtain the enhanced images or videos. Jiang et al.~\cite{Jiang2019} developed EnlightenGAN, with a UNet-based generator and a global-local discriminator. Wang et al.~\cite{Wang2020b} proposed a lightening network that learns the residual between the low-light and normal images with iterative lightening and darkening processes. These methods aim to learn a direct mapping from low-light images to normal images. However, the learning incorporating both luminance enhancement and denoising is challenging, resulting in limited performance under extremely low-light conditions with severe noise levels.

Instead of learning a direct image-to-image mapping, Wang et al.~\cite{Wang2019b} designed a network with intermediate illumination estimation, which is used to brighten the low-light input. Guo et al.~\cite{Guo2020} proposed a zero-reference curve estimation network, which estimates high-order curves for each pixel to adjust the illumination of input image. Ma et al.~\cite{Ma2022} developed a self-calibrated illumination learning framework with cascaded illumination estimation and refinement to achieve fast low-light enhancement. The above methods only enhance the illumination of dark images but cannot suppress noise that is ubiquitous in low-light imaging. 

There are some Retinex-based methods with multiple networks for decomposition and enhancement. Zhang et al.~\cite{Zhang2019} developed a framework that contains a decomposition network, a reflectance restoration network and an illumination adjustment network to deal with the noise and low luminance, respectively. They further improved the restoration network by introducing illumination attention~\cite{Zhang2021c}. Wu et al.~\cite{Wu2022} proposed a deep unfolding framework, with an initialization module for decomposition, an unfolding optimization module to refine the illumination and reflectance iteratively, and an illumination adjustment module to enhance the illumination. Liu et al.~\cite{Liu2021a} proposed an unrolling framework with an illumination estimation module and a noise removal module to perform optimization with lightweight learnable networks. Lu et al.~\cite{Lu2022} developed a framework with two branches, a coefficient estimation branch to predict the coefficients for enhancement, and a joint operation branch to progressively perform joint enhancement and denoising. However, all of the above methods need to train each network or module separately, which leads to a complex training process. In contrast, our method with multiple heads for intermediate task learning can be trained end-to-end.

\subsection{Low-light LF Enhancement}
In addition to low luminance and noise, low-light enhancement for LF images also needs to consider the utilization of multi-view information for better performance. Lamba et al.~\cite{Lamba2021} proposed a network that contains a global representation block to encode angular geometry and a view reconstruction block to restore each view by using multiple neighboring views. They further proposed a three-stage network~\cite{Lamba2022}, with global embedding, view discrimination and RNN-inspired view restoration, to improve the performance. Both networks learn a direct low-light to normal-light mapping. Ge et al.~\cite{Ge2020} used 4D convolution~\cite{Meng2019} with simultaneous spatial and angular feature extraction to construct a LF restoration network. However, 4D convolution leads to a high computational cost and therefore low efficiency. Zhang and Lam~\cite{Zhang2021a} proposed a two-stage framework that includes a multi-to-one network to restore the individual views by fusing the information from other views, and an all-to-all network to refine all the views synchronously with alternate spatial-angular feature extraction. They further improved their work by proposing a Retinex-based framework~\cite{Zhang2021b}, which separates the noise suppression and illumination enhancement, and incorporates interaction and fusion of the spatial and angular information. However, both frameworks are not very efficient, and they were trained on synthetic data with random noise that sometimes deviates from the real noise distribution of LF imaging, resulting in degraded performance when applying on some low-light scenarios. 

\subsection{Vision Transformer}
Vision Transformer (ViT)~\cite{Dosovitskiy2021} is the first report to apply a transformer architecture in a computer vision task. It performs self-attention on the image patches to model the global statistics. However, the global self-attention requires a large computational cost, which grows quadratically with the increase of feature resolution. To tackle this issue, Swin Transformer and its extensions~\cite{Liu2021b,Cao2021,Liu2022} divide the feature maps into non-overlapping windows and calculate self-attention within each window. However, the windows need to be shifted and a lot of layers need to be stacked in order to obtain a global receptive field. Similarly, Uformer~\cite{Wang2022b} adopts non-overlapping window-based self-attention to reduce computation, which also requires a lot of stacked layers for global dependency modeling. To achieve efficient global self-attention, PVT~\cite{Wang2021} and PVTv2~\cite{Wang2022a} employ spatial-reduction attention to merge the tokens of key and value, and Shunted Transformer~\cite{Ren2022} further improves it with multi-scale token merging. However, these methods focus more on the global information without paying much attention to the local detail preservation, which is critical for image restoration task. Therefore, we propose a spatial transformer block with multi-scale self-attention to efficiently encode both the global and fine-grained local information for better restoration.


\section{Low-light restoration transformer for LF images}

\begin{figure}[t]
\centering
\includegraphics[width=0.45\textwidth]{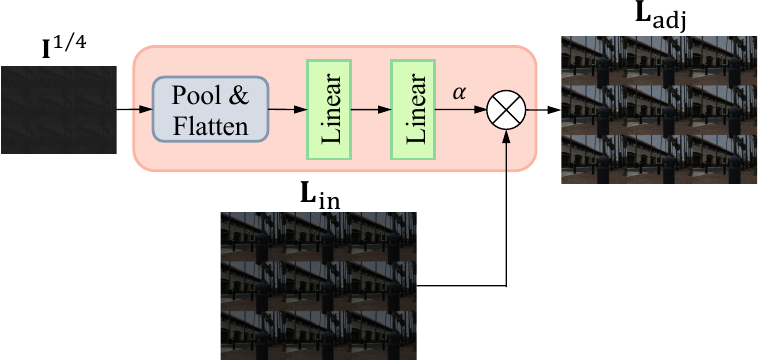} 
\caption{Adaptive ratio adjustment module. It outputs a ratio $\alpha$ by taking the estimated illumination map as input to adaptively adjust the light level of input LF for better detail prediction.}
\label{fig:adaptive module}
\end{figure}

\subsection{Overall Architecture}
The overall architecture of LRT is shown in Fig.~\ref{fig:architecture}, where the left part is for local and global feature extraction, and the right part describes multiple heads to achieve the intermediate tasks. Spatial residual blocks (ResBlocks) are used to extract the local features within each view, and angular transformer blocks are designed to model the dependencies among all the views. In the contracting path, the input degraded LF $\mathbf{L}_\mathrm{in}$ is first fed to the spatial ResBlocks and angular transformer blocks alternately to encode the spatial-angular information. $\frac{1}{2}$-, $\frac{1}{4}$- and $\frac{1}{8}$-scale features are obtained after progressive downsampling. The $\frac{1}{8}$-scale features further pass through several spatial transformer blocks to encode multi-scale local and global information within each view. In order to reduce the computational complexity and memory consumption, the spatial transformer blocks are only applied at the $\frac{1}{8}$ scale to ensure high inference efficiency, since the computational cost grows quadratically with the increase of feature map resolution. During the expanding path, the features are gradually upsampled and fused with the corresponding features in the contracting path by addition. The fused features at different scales are further processed by the spatial ResBlocks and angular transformer blocks, followed by different heads, each of which deals with a specific intermediate task. The intermediate tasks, including denoising, illumination adjustment, refinement and detail enhancement, are performed progressively from $\frac{1}{4}$ scale to full scale by leveraging the extracted features with corresponding scales from the expanding path, which is much more efficient than implementing the whole pipeline in the full scale.

According to the Retinex theory, an image can be expressed as the multiplication of its reflectance and illumination. Hence, the low-light image divided by its illumination derives the normal reflectance that usually suffers from noise. Given this, we introduce an illumination head and a denoising head explicitly performing illumination estimation and noise removal to obtain the normal clean reflectance. The illumination head consisting of a convolution layer with sigmoid activation estimates the illumination of input LF image in the $\frac{1}{4}$-scale branch. The denoising head consisting of a convolution layer with tanh activation is to remove noise from the $\frac{1}{4}$-scale input LF $\mathbf{L}_\mathrm{in}^{1/4}$, which is obtained by downsampling $\mathbf{L}_\mathrm{in}$. The denoised LF $\mathbf{L}_\mathrm{de}^{1/4}$ is divided by the estimated illumination $\mathbf{I}^{1/4}$ in an element-wise manner to yield the $\frac{1}{4}$-scale restored LF $\mathbf{L}_\mathrm{re}^{1/4}$. This configuration separates the denoising and luminance enhancement in order to better handle each one, which can significantly reduce the color distortion of output LF compared to the direct mapping from the low-light noisy image to the normal clean image. These two heads are expressed as
\begin{align}
\mathbf{L}_\mathrm{de}^{1/4}&=\mathbf{L}_\mathrm{in}^{1/4}+\mathbf{R}^{1/4}, \\ 
\mathbf{L}_\mathrm{re}^{1/4}&=\mathbf{L}_\mathrm{de}^{1/4} \oslash \mathbf{I}^{1/4},
\end{align}
where $\mathbf{R}^{1/4}$ is the output of denoising head and $\oslash$ denotes the element-wise division.

$\mathbf{L}_\mathrm{re}^{1/4}$ is further refined in larger scales. The $\frac{1}{2}$-scale branch contains a refinement head that comprises of a convolution layer with tanh activation to output the $\frac{1}{2}$-scale residual map $\mathbf{R}^{1/2}$. $\mathbf{L}_\mathrm{re}^{1/4}$ is upsampled and added to $\mathbf{R}^{1/2}$ to yield the $\frac{1}{2}$-scale restored LF $\mathbf{L}_\mathrm{re}^{1/2}$, with
\begin{align}
\mathbf{L}_\mathrm{re}^{1/2}&=f_\mathrm{up}(\mathbf{L}_\mathrm{re}^{1/4})+\mathbf{R}^{1/2},
\end{align} 
where $f_\mathrm{up}(\cdot)$ denotes the $2\times$ upsampling operation.

The full-scale branch also has a refinement head to output the residual map $\mathbf{R}$. Similarly, $\mathbf{L}_\mathrm{re}^{1/2}$ is further upsampled and added to $\mathbf{R}$ to obtain the full-scale restored LF $\mathbf{L}_\mathrm{re}$. In addition, it has a high-frequency head to predict the high-frequency components to enhance the local details. Considering the extremely low-light inputs with nearly invisible high-frequency details, we introduce an adaptive ratio adjustment module to adjust the light level of input LF adaptively. As shown in Fig.~\ref{fig:adaptive module}, it takes the estimated illumination map $\mathbf{I}^{1/4}$ as input to serve as a luminance clue. $\mathbf{I}^{1/4}$ is first pooled and flattened and then fed to two linear layers to output a factor $\alpha$. $\mathbf{L}_\mathrm{in}$ multiplied by $\alpha$ obtains the adjusted LF $\mathbf{L}_\mathrm{adj}$. The concatenation of $\mathbf{L}_\mathrm{adj}$ and the full-scale features passes through the high-frequency head to output the high-frequency map $\mathbf{H}$. This adaptive module aims to find a proper ratio to ensure that $\mathbf{L}_\mathrm{adj}$ can provide clear cues for predicting the local details. The addition of $\mathbf{H}$ and $\mathbf{L}_\mathrm{re}$ derives the final output LF image $\mathbf{L}_\mathrm{out}$, with
\begin{align}  
\mathbf{L}_\mathrm{re}&=f_\mathrm{up}(\mathbf{L}_\mathrm{re}^{1/2})+\mathbf{R}, \\
\mathbf{L}_\mathrm{out}&=\mathbf{L}_\mathrm{re}+\mathbf{H}.
\end{align}

\begin{figure}[t]
\centering
\includegraphics[width=0.48\textwidth]{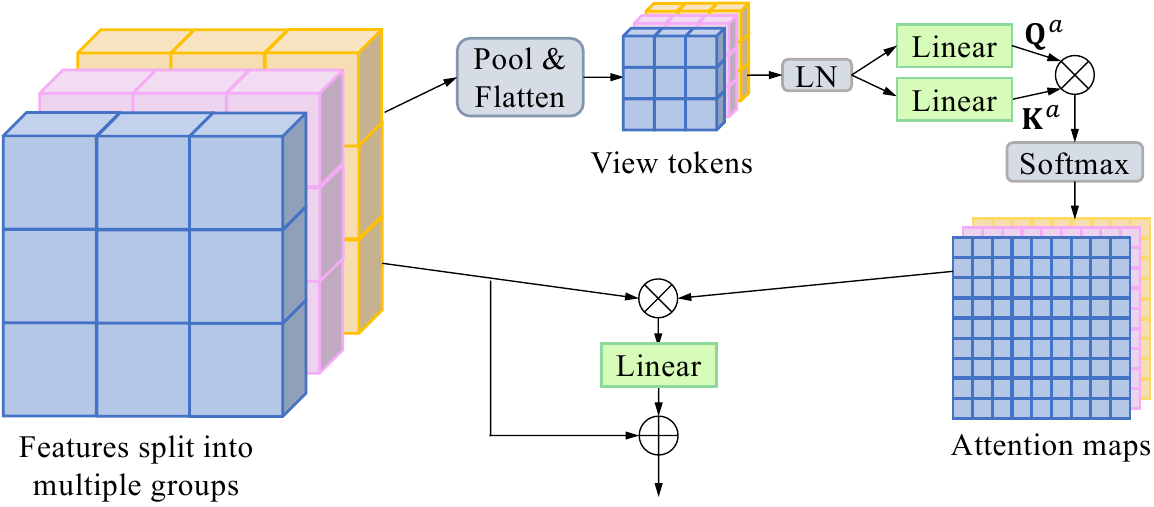} 
\caption{Angular transformer block. View tokens are obtained by applying pooling and flatten operations to the view features, and they are further processed by linear layers to generate the query and key. The view features are used as the value. The query, key and value are split into different groups to compute self-attention.}
\label{fig:angular transformer}
\end{figure}

\begin{figure*}[t]
\centering
\includegraphics[width=1\textwidth]{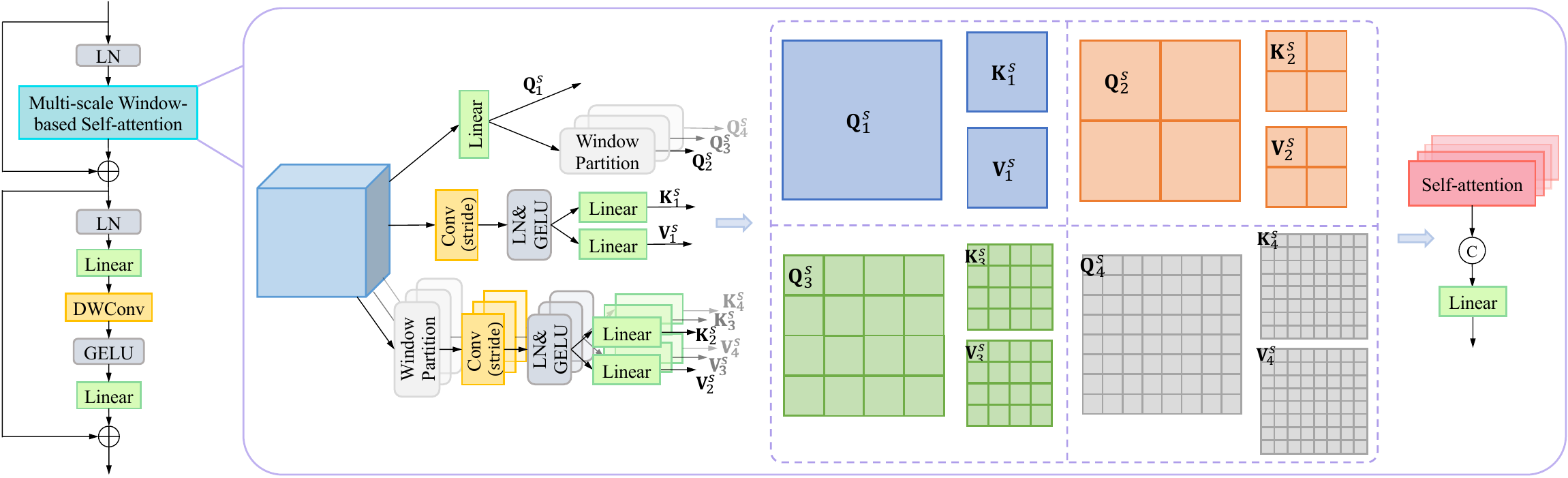} 
\caption{Spatial transformer block. It adopts multi-scale window-based self-attention with four groups to encode both the global and local features. The first group computes the global self-attention, and the other three groups compute self-attention within the local windows with different sizes. Stride convolutions are used to reduce the lengths of keys and values to improve efficiency.}
\label{fig:spatial transformer}
\end{figure*}

\subsection{Angular Transformer Block}
The angular transformer block aims to explore the global angular dependencies among all the views. Fig.~\ref{fig:angular transformer} shows its structure (taking $3\times 3$ view input as an example). The feature $\mathbf{F}\in \mathbb {R}^{u\times v\times c\times h\times w}$ (with angular resolution $u\times v$, spatial resolution $h\times w$, and channel number $c$) is first split into $m$ groups along the channel dimension, with $\{\mathbf{F}_i\}_{i=1}^m$ ($\mathbf{F}_i\in \mathbb {R}^{u\times v\times \frac{c}{m}\times h\times w}$). There would be a huge computational cost if all the feature elements of each view are used to calculate the attention. Thus, for each group, we apply the pooling and flattening operations to the view features to obtain the view tokens, each of which incorporates the core information of one view. Then, the view tokens pass through a layer normalization (LN) and two linear layers to generate the query $\mathbf{Q}^a_i\in \mathbb{R}^{uv\times d}$ and key $\mathbf{K}^a_i\in \mathbb{R}^{uv\times d}$, with dimension $d$. $\mathbf{F}_i$ is used as the value. The procedures of calculating the angular self-attention for group $i$ is given by
\begin{align}
\mathbf{Q}^a_i&=f_\mathrm{LN}\big(f_\mathrm{pool}(\mathbf{F}_i)\big)\mathbf{W}^Q_i, \\ 
\mathbf{K}^a_i&=f_\mathrm{LN}\big(f_\mathrm{pool}(\mathbf{F}_i)\big)\mathbf{W}^K_i, \\  
\mathbf{G}^a_i&=\mathrm{softmax}\Big(\frac{\mathbf{Q}^a_i(\mathbf{K}^a_i)^T}{\sqrt{d}}\Big)\mathbf{F}_i,
\end{align} 
where $f_\mathrm{pool}(\cdot)$ means pooling with flattening operated on each view feature, $f_\mathrm{LN}(\cdot)$ denotes the layer normalization, and $\mathbf{W}^Q_i$ and $\mathbf{W}^K_i$ are the parameter matrices of the linear layers.

The outputs of all the groups are concatenated, and then fed to another linear layer to fuse the features of different groups, whose output is added to the input feature to obtain the output feature $\mathbf{F}'$ with cross-view information. This process is expressed as
\begin{align}
\mathbf{F}'=[\mathbf{G}^a_1,\mathbf{G}^a_2,\cdots,\mathbf{G}^a_m]\mathbf{W}^G+\mathbf{F},
\end{align}
where $[\cdot]$ denotes the concatenation operation and $\mathbf{W}_G$ denotes the parameter matrix of the linear layer for fusion.

Given the feature with size $u\times v\times c\times h\times w$, the complexity of our angular transformer block $\Omega_A$ is calculated as 
\begin{align}
\Omega_A=uvc^2\frac{p^2+2}{m}+(uv)^2hwc+(uv)^2c+uvhwc^2,
\end{align}
where $p$ is the size of view feature after pooling. Compared to~\cite{Liang2022}, which computes angular self-attention within each macro-pixel with the complexity $4uvhwc^2+2(uv)^2hwc$, our angular transformer block involves a lower computational cost due to the efficient view-token scheme. 

In this way, each view can integrate the complementary information from all the other views according to the attention maps. Moreover, it enables synchronous restoration for all the views in each forward process. 

\subsection{Spatial Transformer Block}
The common transformer architecture performs global self-attention by computing the correlation among all the tokens, which results in high computational complexity, especially for the dense prediction tasks with high-resolution feature maps. Self-attention within local windows is a promising solution for reducing the computation burden. However, it cannot model the global dependencies and the relationships across different windows effectively.

For efficient modeling on both the global and local dependencies within each view, we propose a spatial transformer block (Fig.~\ref{fig:spatial transformer}) with the multi-scale window-based self-attention, which contains four groups to encode features in different scales. The first group (global group) computes self-attention within the overall feature maps, and the other three groups (local groups) compute self-attention within the local windows with different sizes. The $\frac{1}{8}$-scale feature after LN passes through a linear layer to obtain the query $\mathbf{Q}_1^s$ for the global group, and $\mathbf{Q}_2^s$, $\mathbf{Q}_3^s$ and $\mathbf{Q}_4^s$ for the local groups, which are partitioned into $2\times2$, $4\times4$ and $8\times8$ windows, respectively. To further reduce the computational cost, we decrease the lengths of keys and values using convolution layers with large strides to downsample the features. Then, LN, GELU non-linearity~\cite{Hendrycks2016} and linear layers are employed to generate $\{\mathbf{K}_j^s\}_{j=1}^4$ and $\{\mathbf{V}_j^s\}_{j=1}^4$. For the local groups, stride convolutions are applied within their respective local windows. The spatial self-attention of each group is calculated by
\begin{align}
\mathbf{G}^s_j=\mathrm{softmax}\Big(\frac{\mathbf{Q}^s_j(\mathbf{K}^s_j)^T}{\sqrt{d}}\Big)\mathbf{V}^s_j.
\end{align}
The outputs of all the groups are concatenated and then fed to a linear layer for fusion. 

Given the spatial feature with size $c\times h\times w$, the kernel size of stride convolution $t$ and the number of windows $n\times n$ for each group, the complexity of our spatial self-attention $\Omega_S$ with four groups $\mathcal{G}=\{(1,4),(2,4),(4,2),(8,2)\}$ is calculated as 
\begin{align}
\Omega_S&=\sum_{\substack{(n,t)\in \mathcal{G}}}\Big(\frac{hwc^2}{2n^2}+\frac{hwc^2}{4(nt)^2}+\frac{(hw)^2c}{2(nt)^2n^2}\Big)n^2+2hwc^2 \nonumber\\ 
&=4hwc^2+\frac{5}{32}hwc^2+\frac{25}{512}(hw)^2c,
\end{align}
which is much less than the complexity of common global self-attention, with $4hwc^2+2(hw)^2c$.

After the self-attention layer, the features pass through the feed-forward layers. Similar to PVTv2~\cite{Wang2022a}, we introduce a depth-wise convolution layer between the two linear layers to complement the local information.  

\subsection{Loss Function}
In our LRT, each head corresponds to specific loss terms. For the $\frac{1}{4}$-scale denoising head, the denoising loss is
\begin{align}
\ell_\mathrm{de}=\|\mathbf{L}_\mathrm{de}^{1/4}-\mathbf{L}_\mathrm{low}^{1/4}\|_1,
\end{align}
where $\mathbf{L}_\mathrm{low}^{1/4}$ is the $\frac{1}{4}$-scale clean low-light LF, with the same illumination as the input LF and obtained from the ground-truth $\mathbf{L}_\mathrm{gt}$, and $\|\cdot\|_1$ is the $\ell_1$ norm.

For the $\frac{1}{4}$-scale illumination head, the estimated illumination map is expected to be smooth while preserving the object structures. Thus, the smoothness loss is introduced, with
\begin{align}
\ell_\mathrm{sm}=|\nabla \mathbf{I}^{1/4}|\times e^{(-\eta |\nabla \mathbf{L}_\mathrm{gt}^{1/4}|)},
\end{align}
where $\nabla$ denotes the gradients along both the horizontal and vertical directions, $\eta$ is a hyper-parameter to adjust the structure awareness, and $\mathbf{L}_\mathrm{gt}^{1/4}$ is the $\frac{1}{4}$-scale ground-truth LF serving as a structure reference.

For better structure and contrast preservation, we propose an illumination reference loss by utilizing the Y channel (YUV color space) of low-light LF, defined as
\begin{align}
\ell_\mathrm{ref}=\|f_\mathrm{nor}(\mathbf{I}^{1/4})-f_\mathrm{nor}(\mathbf{Y}^{1/4})\|_1,
\end{align} 
where $\mathbf{Y}^{1/4}$ is the Y channel of $\mathbf{L}_\mathrm{low}^{1/4}$, $f_\mathrm{nor}(\cdot)$ is to normalize the illumination map and Y channel to $[0,1]$, with $f_\mathrm{nor}(\mathbf{I}^{1/4})=\frac{\mathbf{I}^{1/4}-\mathrm{min}(\mathbf{I}^{1/4})}{\mathrm{max}(\mathbf{I}^{1/4})-\mathrm{min}(\mathbf{I}^{1/4})}$. By introducing this constraint, the illumination head can learn the light distribution of objects more effectively with the guidance of the Y channel. 

The high-frequency map is learned through the supervision of the ground truth $\mathbf{F}_\mathrm{gt}$, which is obtained by
\begin{align}
\mathbf{F}_\mathrm{gt}=\mathbf{L}_\mathrm{gt}-f_\mathrm{gau}(\mathbf{L}_\mathrm{gt}),
\end{align}
where $f_\mathrm{gau}(\cdot)$ denotes the Gaussian filter. The high-frequency loss is written as
\begin{align}
\ell_\mathrm{hf}=\|\mathbf{F}-\mathbf{F}_\mathrm{gt}\|_1.
\end{align}

The reconstruction loss is applied to all the intermediate restored LFs and the final output LF, expressed as
\begin{align}
\ell_\mathrm{rec}&=\|\mathbf{L}_\mathrm{re}^{1/4}-\mathbf{L}_\mathrm{gt}^{1/4}\|_1+\|\mathbf{L}_\mathrm{re}^{1/2}-\mathbf{L}_\mathrm{gt}^{1/2}\|_1 \nonumber\\
&\quad+\|\mathbf{L}_\mathrm{re}-\mathbf{L}_\mathrm{gt}\|_1+\|\mathbf{L}_\mathrm{out}-\mathbf{L}_\mathrm{gt}\|_1.
\end{align} 

In addition, we impose the structural similarity (SSIM)\cite{Wang2004} loss to the final output LF to improve the visual quality, with
\begin{align}
\ell_\mathrm{SSIM}=1-\mathrm{SSIM}(\mathbf{L}_\mathrm{out},\mathbf{L}_\mathrm{gt}).
\end{align}

Thus, the full loss for training LRT is the combination of the above loss terms with their respective coefficients, written as
\begin{align}\label{eq:full loss}
\ell_\mathrm{full}&=\lambda_\mathrm{de}\ell_\mathrm{de}+\lambda_\mathrm{rec}\ell_\mathrm{rec}+\lambda_\mathrm{SSIM}\ell_\mathrm{SSIM}\nonumber \\
&\quad+\lambda_\mathrm{sm}\ell_\mathrm{sm}+\lambda_\mathrm{ref}\ell_\mathrm{ref}+\lambda_\mathrm{hf}\ell_\mathrm{hf}.
\end{align}

\section{Experiments}

\begin{figure*}[t]
\centering
\subfloat[]{\includegraphics[height=0.14\linewidth]{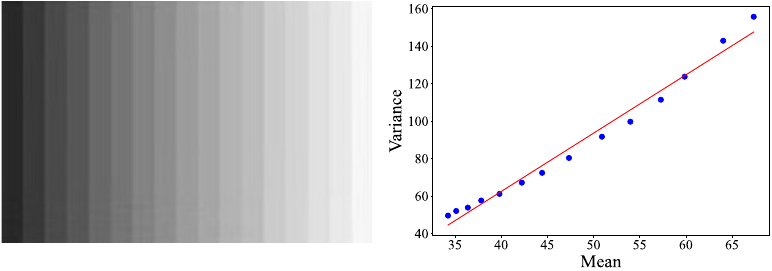}}
\hfil
\subfloat[]{\includegraphics[height=0.14\linewidth]{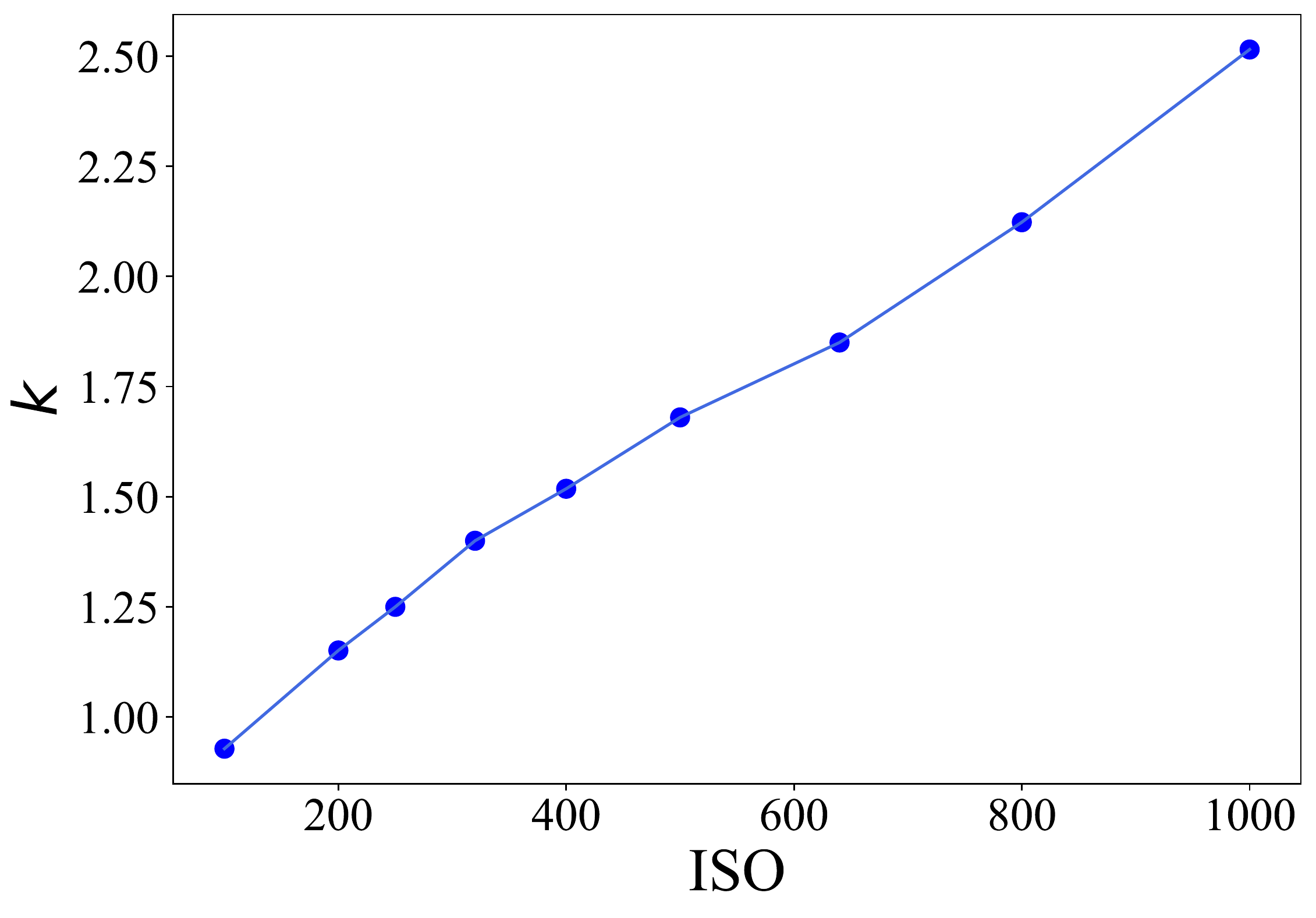}}
\hfil
\subfloat[]{\includegraphics[height=0.145\linewidth]{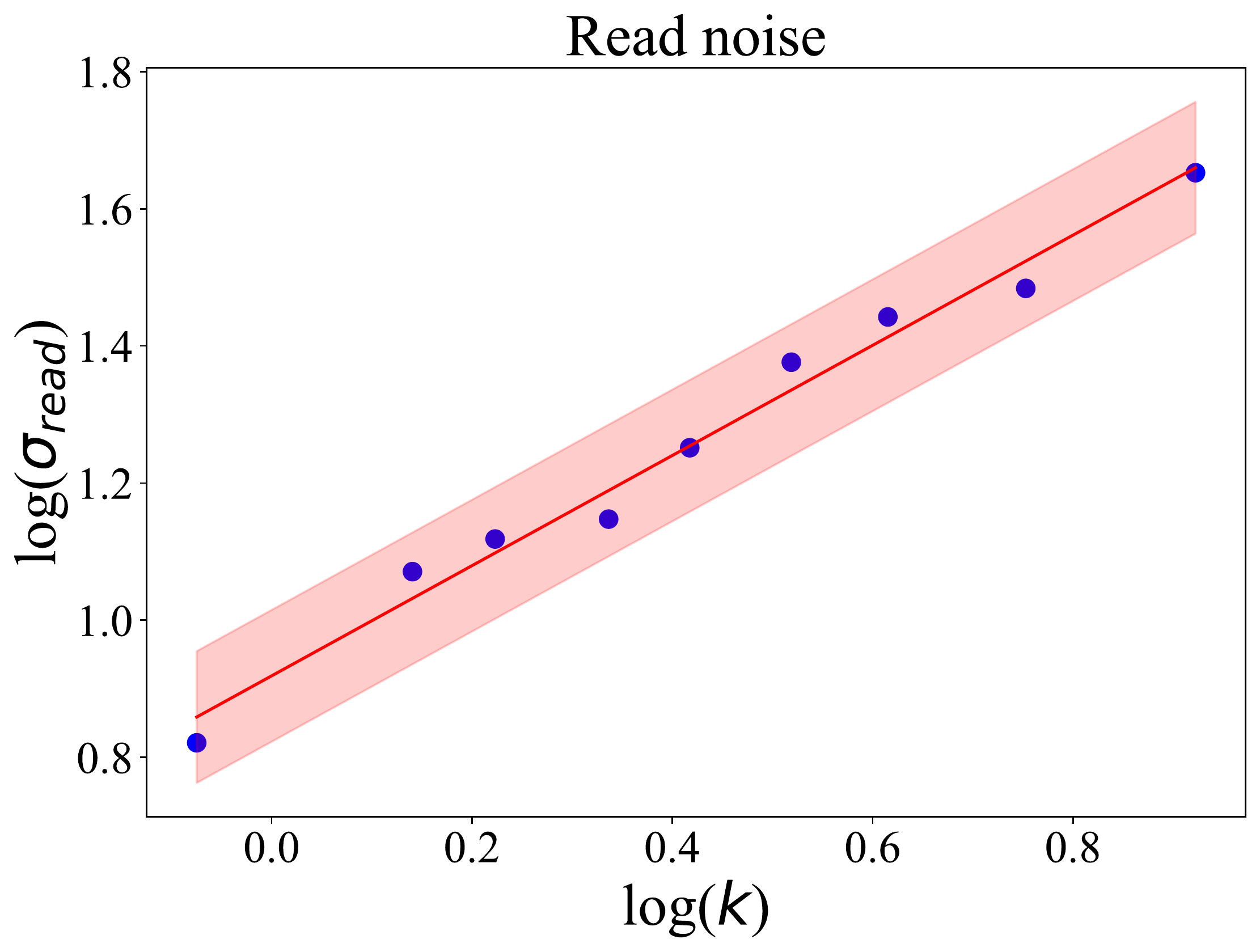}}
\hfil
\subfloat[]{\includegraphics[height=0.145\linewidth]{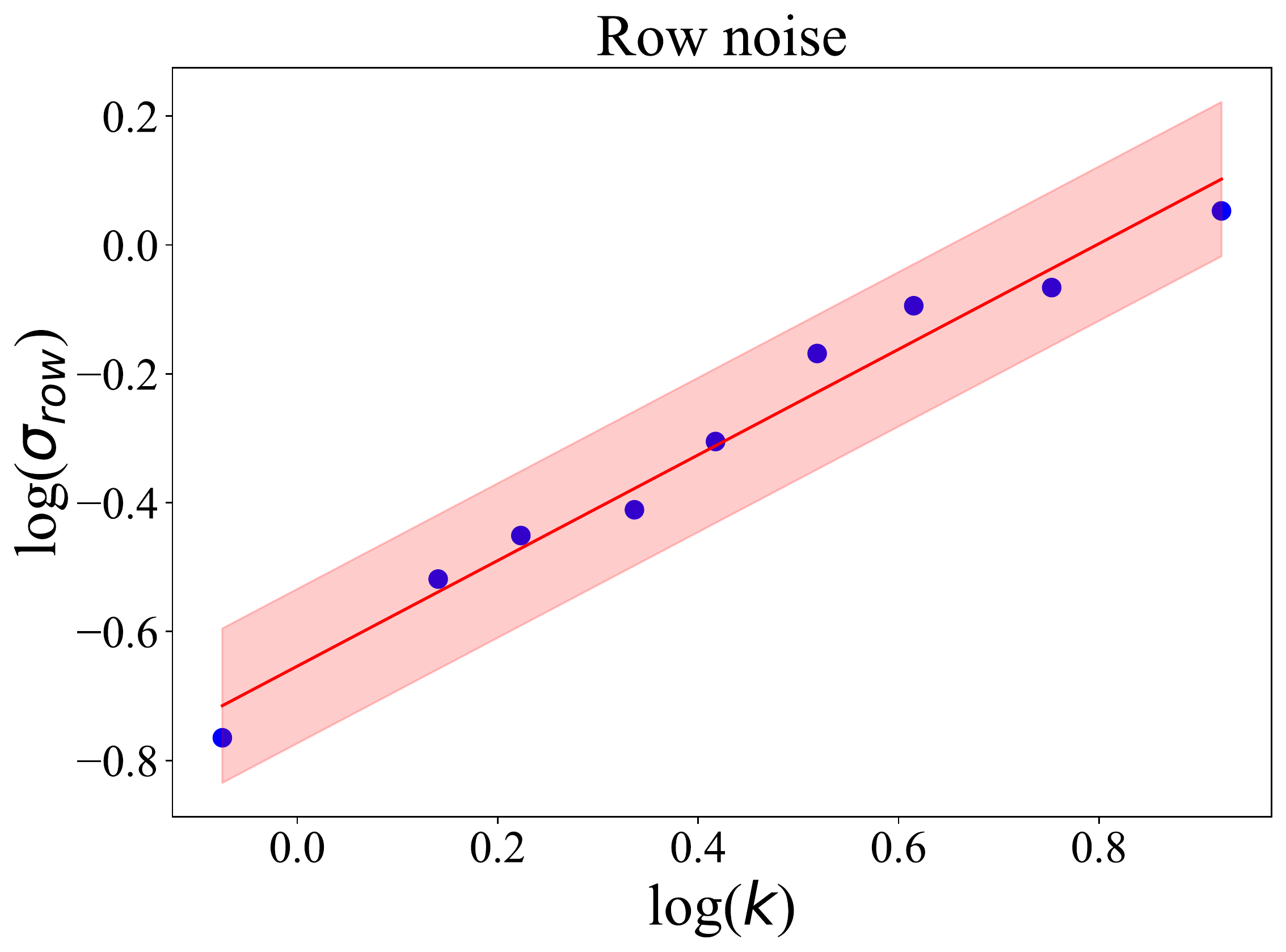}}
\caption{(a) Estimate $k$ under a specific ISO with the gray-scale images and linear regression. (b) $k$ values under different ISOs. (c) Joint distribution of $k$ and $\sigma_\mathrm{read}$. (d) Joint distribution of $k$ and $\sigma_\mathrm{row}$.}
\label{fig:estimate noise}
\end{figure*}

\subsection{Synthesis of Dark LF Images}
\subsubsection{Noise parameter estimation} To synthesize more realistic dark LF images, we first estimated the noise parameters of a LF camera. As with the conventional camera, the imaging process of the LF camera also introduces a variety of noise sources. 

The number of photoelectrons $\mathbf{E}_\mathrm{photon}$ follows a Poisson distribution $\mathcal{P}$ and is proportional to the exposure time $\tau$, luminous flux $\Phi$, and quantum efficiency $\alpha$, with $\mathbf{E}_\mathrm{photon}\sim \mathcal{P}(\tau\alpha\Phi)$, which is the source of shot noise. The generated electrons in the lightless environment also follows a Poisson distribution, which introduces the dark noise, with $\mathbf{E}_\mathrm{dark}\sim \mathcal{P}(\tau \mathbf{D})$ ($\mathbf{D}$ denotes the electrons produced per unit time under current temperature). Therefore, the total number of electrons before the amplifier is written as $\mathbf{E}_\mathrm{total}\sim \mathcal{P}(\tau \alpha\Phi+\tau \mathbf{D})$ $(\mathbf{E}_\mathrm{total}=\mathbf{E}_\mathrm{photon}+\mathbf{E}_\mathrm{dark})$.

The collected electrons are converted to the voltage, and then passed through the analog amplifier and analog-to-digital converter (ADC) to obtain the pixel value. These stages introduce the read noise $\mathbf{N}_\mathrm{read}$ following a Gaussian distribution $\mathcal{N}(0,\sigma_\mathrm{read})$, and the quantization noise $\mathbf{N}_\mathrm{quan}$ following a uniform distribution $\mathcal{U}(-\frac{q}{2},\frac{q}{2})$ with the quantization step $q$. Furthermore, banding noise usually exists under low-light imaging. Here, we mainly consider the row noise $\mathbf{N}_\mathrm{row}$, which is modeled as an offset added to each row and sampled from a Gaussian distribution $\mathcal{N}(0,\sigma_\mathrm{row})$. Thus, the raw plenoptic image from a LF camera can be expressed as
\begin{align}
\mathbf{L}=k\mathbf{E}_\mathrm{total}+\mathbf{N}_\mathrm{read}+\mathbf{N}_\mathrm{quan}+\mathbf{N}_\mathrm{row},
\end{align}
where $k$ is the system gain relevant to ISO. 

Let $\mathbf{N}_\mathrm{add}=\mathbf{N}_\mathrm{read}+\mathbf{N}_\mathrm{quan}+\mathbf{N}_\mathrm{row}$. The mean and variance of the raw data are given by
\begin{align}
\mathbb{E}(\mathbf{L})&=k(\tau\alpha\Phi+\tau \mathbf{D}) \label{eq: mean}\\ 
\mathrm{Var}(\mathbf{L})&=k^2(\tau\alpha\Phi+\tau \mathbf{D})+\mathrm{Var}(\mathbf{N}_\mathrm{add}). 
\end{align}
Then, we have
\begin{align}
\mathrm{Var}(\mathbf{L})=k\mathbb{E}(\mathbf{L})+\mathrm{Var}(\mathbf{N}_\mathrm{add}).
\end{align}

To estimate $k$ under a specific ISO, we capture a series of grayscale images (Fig.~\ref{fig:estimate noise}(a)) by the LF camera. The raw plenoptic images are rectified by the estimated centroids of the micro images to obtain the aligned images. The mean and variance within a small region of each gray scale on the aligned image are calculated. Then, we apply linear regression to find an optimal line fitting the mean and variance, and the slope of the line is the estimated $k$. 

To estimate the signal-independent noise, we capture a series of dark plenoptic images under each ISO. From Eq.~\ref{eq: mean}, we have $\mathbb{E}(\mathbf{L})=k\tau \mathbf{D}$ when $\Phi=0$. Thus, we can calculate the mean of these dark images under a specific ISO and exposure time, and the mean divided by $k$ can be the estimated dark noise. For the row noise, the mean values of each row can be treated as the row noise intensities~\cite{Wei2021} since the mean of the other noise sources approximates to $0$. Then, the standard deviation $\sigma_\mathrm{row}$ can be estimated by fitting a normal distribution. After removing the dark noise and row noise from the dark image, we can estimate the standard deviation of read noise $\sigma_\mathrm{read}$ by fitting a normal distribution.

\begin{figure*}[t]
\centering
\includegraphics[width=1\textwidth]{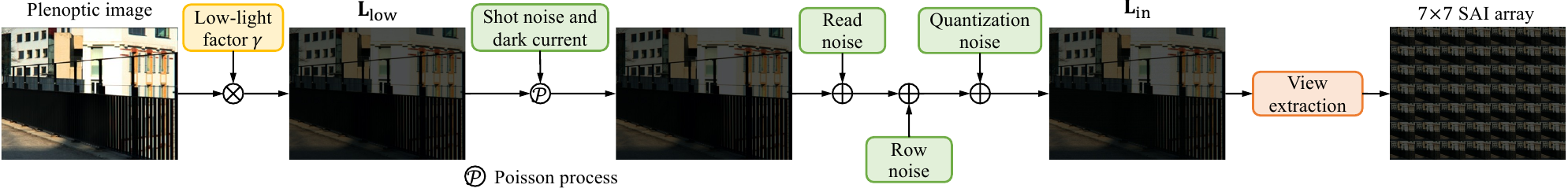} 
\caption{The plenoptic image is first multiplied by a low-light factor, and then a variety of noise sources are added to obtain the low-light noisy plenoptic image, which is converted to the SAI array with $7\times 7$ views.}
\label{fig:pipeline}
\end{figure*}

\begin{figure}[!hbt]
\centering
\includegraphics[width=0.5\textwidth]{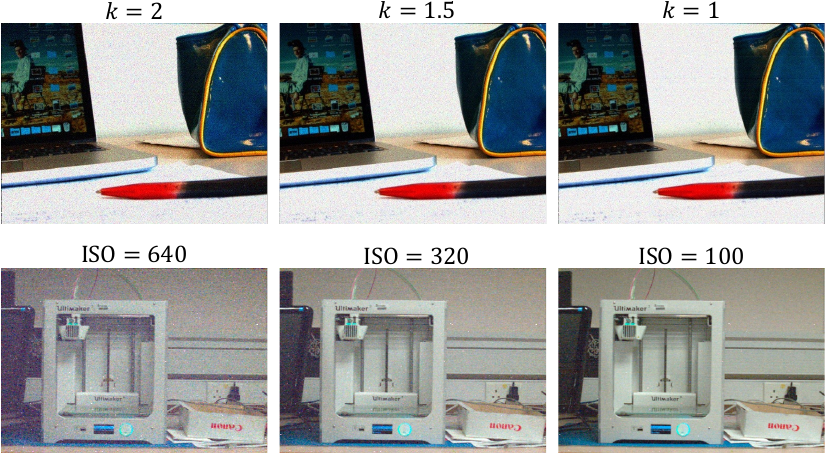} 
\caption{The first row presents the central views of the synthetic LF images with different $k$ values. The second row shows the central views of the captured LF images under different ISOs.}
\label{fig:example}
\end{figure}

With these methods, we estimated $k$, $\sigma_\mathrm{read}$ and $\sigma_\mathrm{row}$ under different ISOs for the LF camera. Fig.~\ref{fig:estimate noise}(b) plots the estimated $k$ under ISOs ranging from $100$ to $1000$. Following~\cite{Wei2021}, we calculated $\log$ values of the estimated parameters and apply linear regression to obtain the relationship of $(\log{k},\log{\sigma_\mathrm{read}})$ and $(\log{k},\log{\sigma_\mathrm{row}})$, as shown in Fig.~\ref{fig:estimate noise}(c) and (d). The noise parameters can be sampled from the red shadow regions, which are derived according to the standard deviation of linear fitting. 

\subsubsection{Synthesis pipeline} 
To synthesize dark noisy LF images, clean images from Kalantari~\cite{Kalantari2016}, Stanford~\cite{Raj2016} and EPFL~\cite{Rerabek2016} datasets are chosen as the ground truths. A low-light factor $\beta$ is first sampled uniformly from $[0.05,0.2]$ and multiplied to the ground truth to obtain the low-light clean LF $\mathbf{L}_\mathrm{low}$. $k$ is also sampled uniformly from its estimated range. $\sigma_\mathrm{read}$ and $\sigma_\mathrm{row}$ are derived according to their modeled relationships with $k$. With the estimated noise parameters $\hat{\sigma}_\mathrm{read}$ and $\hat{\sigma}_\mathrm{row}$, dark current $\hat{\mathbf{E}}_\mathrm{dark}$, and the fixed quantization step $q$, we can add these major noise sources to $\mathbf{L}_\mathrm{low}$ to obtain the low-light noisy LF $\mathbf{L}_\mathrm{in}$. The synthesis pipeline is formulated as
\begin{align}
\mathbf{L}_\mathrm{low}&=\beta \mathbf{L}_\mathrm{gt}, \\ 
\mathbf{L}_\mathrm{in}&=k\mathcal{P}\left(\frac{\mathbf{L}_\mathrm{low}}{k}+\hat{\mathbf{E}}_\mathrm{dark}\right)+\mathcal{N}(0,\hat{\sigma}_\mathrm{read}) \nonumber\\ 
&\quad+\mathcal{N}(0,\hat{\sigma}_\mathrm{row})+\mathcal{U}(-\frac{q}{2},\frac{q}{2}).
\end{align}

The synthesis procedures are implemented on the plenoptic image, as shown in Fig.~\ref{fig:pipeline}, and the noise and low luminance are allocated to different views after converting the plenoptic image to the SAI array. We preserve only the central $7\times 7$ views that have very similar illumination and noise levels. Fig.~\ref{fig:example} presents the central views of our synthetic LF images with different $k$ values and the captured LF images under different ISOs, which are brightened to clearly show the noise distributions. We use these synthetic LF images to train our network, which alleviates the need to capture a large number of low-light/normal image pairs.

\subsection{Implementation Details}
During each training step, the input LF images were synthesized from the ground truths with the randomly selected low-light factors and noise parameters, and cropped to $256\times 256$ patches randomly. The $\frac{1}{2}$- and $\frac{1}{4}$-scale images were obtained by downsampling the full-scale images. Gaussian smoothing was applied before each downsampling operation to avoid the aliasing effect~\cite{Zhang2021d}. Our LRT was trained end-to-end using the loss in Eq.~\ref{eq:full loss} for about $300$ epochs, with coefficients $\lambda_\mathrm{de}=10$, $\lambda_\mathrm{rec}=5$, $\lambda_\mathrm{SSIM}=1$, $\lambda_\mathrm{sm}=0.1$, $\lambda_\mathrm{ref}=1$, and $\lambda_\mathrm{hf}=1$. Adam optimizer was used with learning rate set to $5\times 10^{-4}$ initially and decayed by multiplying $0.8$ after every $50$ epochs. Our LRT was implemented using PyTorch on the NVIDIA Tesla P100 GPU.

\subsection{Comparison}
We compared our method with several state-of-the-art low-light enhancement methods for the single images, including DeepUPE~\cite{Wang2019b}, Zero-DCE~\cite{Guo2020}, RUAS~\cite{Liu2021a}, KinD++~\cite{Zhang2021c} and URetinex-Net~\cite{Wu2022}, and for the LF images, including LFRetinex~\cite{Zhang2021b}, L3Fnet~\cite{Lamba2021} and TSNet~\cite{Lamba2022}. The single image methods restore each view separately without using the information of other views, while the LF image methods leverage multi-view information to restore each view. We trained all these methods on our synthetic dataset and evaluated them on the dark LF images captured by us to show their performance on the practical low-light restoration. PSNR, SSIM and LPIPS~\cite{Zhang2018} are the evaluation metrics. 

\begin{figure*}[!hbt]
\centering
\includegraphics[width=1\textwidth]{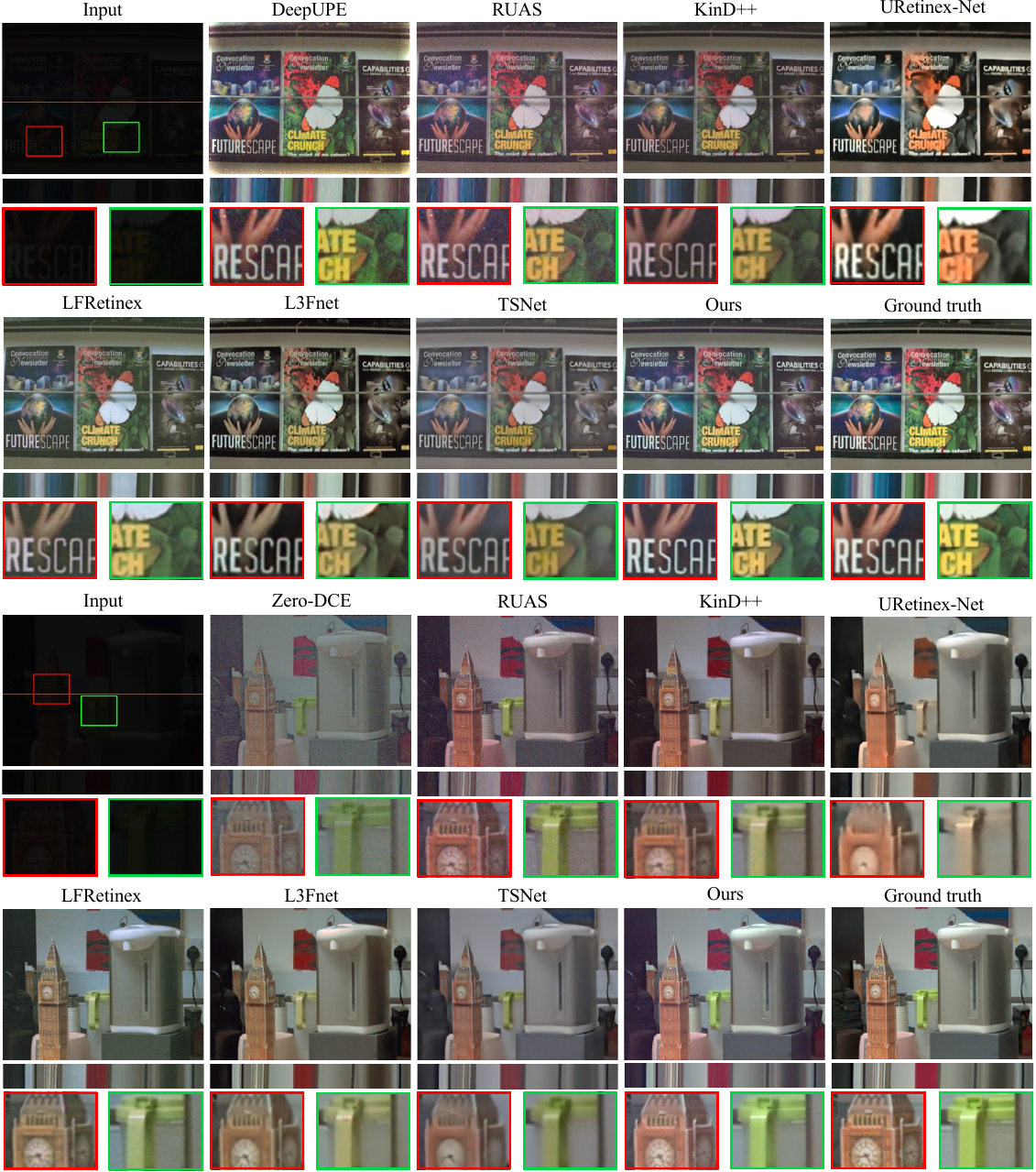} 
\caption{Visual results of different methods on our captured low-light LF images with $\mathrm{ISO}=200$ and $\mathrm{ISO}=500$. The central views and EPIs are presented, with zoomed patches to show the local details.}
\label{fig:capture_com}
\end{figure*}

\begin{figure*}[!hbt]
\centering
\includegraphics[width=1\textwidth]{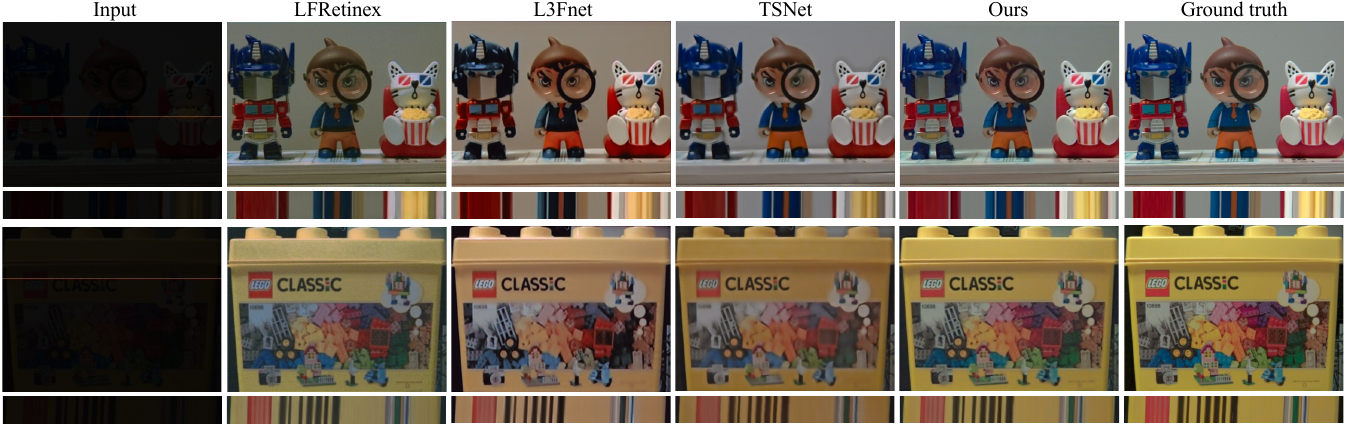} 
\caption{Visual results of different LF image methods on our captured dark LF images with $\mathrm{ISO}=100$ and $200$.}
\label{fig:LF_com}
\end{figure*}

\begin{figure*}[!hbt]
\centering
\includegraphics[width=1\textwidth]{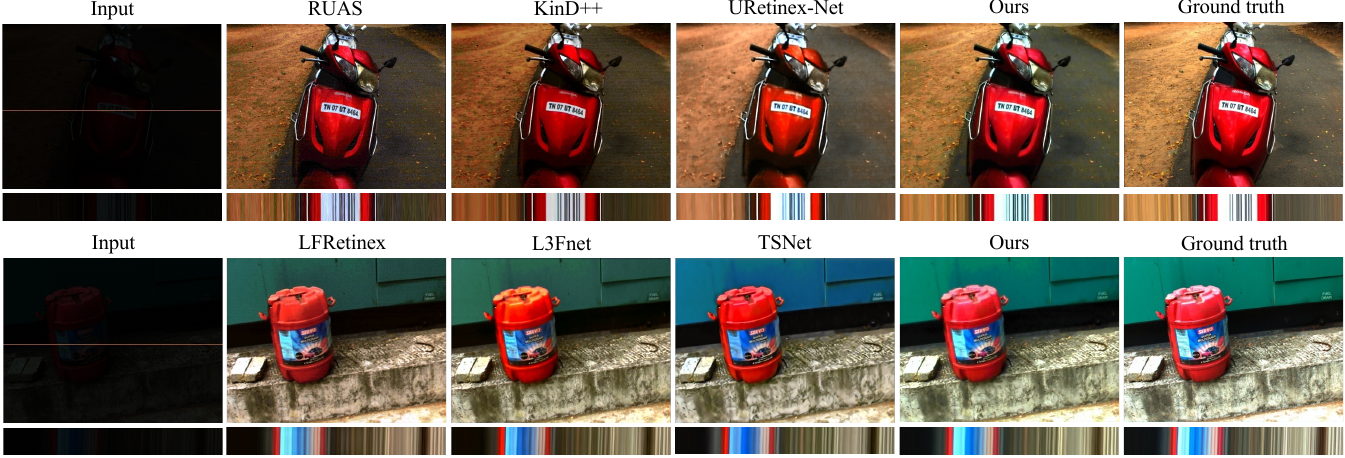} 
\caption{Visual results of different methods on the LF images from L3F dataset~\cite{Lamba2021}. The central view and EPIs are presented.}
\label{fig:L3F_com}
\end{figure*}

\subsubsection{Evaluation on our captured dark LF images}

\begin{table}[t]
\caption{Quantitative results on our captured LF images. Bold: Best.}
\centering
\begin{tabular}{c|ccc}
\toprule
Method &  PSNR$\uparrow$ & SSIM$\uparrow$ & LPIPS$\downarrow$ \\[0.5ex]
\midrule
\textbf{Single image method} & & & \\
DeepUPE~\cite{Wang2019b}     & 20.17 & 0.705 & 0.166 \\
Zero-DCE~\cite{Guo2020}      & 18.73 & 0.693 & 0.205 \\
RUAS~\cite{Liu2021a}         & 21.54 & 0.730 & 0.165 \\
KinD++~\cite{Zhang2021c}     & 21.82 & 0.795 & 0.133 \\
URetinex-Net~\cite{Wu2022}   & 23.07 & 0.808 & 0.191 \\
\midrule
\textbf{LF image method}     & & &\\
LFRetinex~\cite{Zhang2021b}  & 22.16 & 0.826 & 0.104 \\
L3Fnet~\cite{Lamba2021}      & 23.13 & 0.825 & 0.137 \\
TSNet~\cite{Lamba2022}       & 22.88 & 0.813 & 0.150 \\
Ours                         & \textbf{26.67} & \textbf{0.857} & \textbf{0.086} \\
\bottomrule
\end{tabular}
\label{tab:capture results}
\end{table}

We captured some LF images in the low-light environment with ISO ranging from $100$ to $640$ to evaluate different methods. The quantitative results are listed in Table~\ref{tab:capture results}, where we observe that our method outperforms the other single image and LF image methods, with higher PSNR and SSIM, and lower LPIPS. 

The visual results under $\mathrm{ISO}=200$ and $500$ are shown in Fig.~\ref{fig:capture_com}, including the central views, epipolar-plane images (EPIs) and zoomed patches. It can be seen that DeepUPE and Zero-DCE can only enhance illumination without suppressing noise since their models do not incorporate a denoising module. RUAS adopts the unfolding optimization approach to suppress noise, which is not very effective when there is serious noise, and therefore results in residual noise in the restored LF images. Moreover, the images restored by the Retinex-based methods (KinD++, URetinex-Net and LFRetinex) and the direct mapping methods (L3Fnet and TSNet) all suffer obvious color distortion compared to the ground truths. In contrast, our method can achieve better luminance enhancement and noise suppression with little color distortion, obtaining LF images with more compelling visual qualities. 

Fig.~\ref{fig:LF_com} gives further comparison of different LF image methods, which suggests that our method achieves better color recovery to obtain images with closer color distribution to the ground truths, and restores more and clearer high-frequency details compared to the other LF images methods.

\subsubsection{Evaluation on the L3F dataset}

\begin{table}[t]
\caption{Quantitative results on the L3F dataset. Bold: Best.}
\centering
\begin{tabular}{c|ccc}
\toprule
Method  & PSNR$\uparrow$ & SSIM$\uparrow$ & LPIPS$\downarrow$ \\[0.5ex]
\midrule
\textbf{Single image method} & & & \\
DeepUPE~\cite{Wang2019b}     & 19.94 & 0.596 & 0.279 \\
Zero-DCE~\cite{Guo2020}      & 19.12 & 0.574 & 0.325 \\
RUAS~\cite{Liu2021a}         & 20.03 & 0.626 & 0.248 \\
KinD++~\cite{Zhang2021c}     & 20.36 & 0.668 & 0.182 \\
URetinex-Net~\cite{Wu2022}   & 21.55 & 0.704 & 0.205 \\
\midrule
\textbf{LF image method}     & & & \\
LFRetinex~\cite{Zhang2021b}  & 21.00 & 0.693 & 0.186 \\
L3Fnet~\cite{Lamba2021}      & 22.01 & 0.755 & 0.149 \\
TSNet~\cite{Lamba2022}       & 22.51 & 0.733 & 0.163 \\
Ours                         & \textbf{24.31} & \textbf{0.801} & \textbf{0.119} \\
\bottomrule
\end{tabular}
\label{tab:L3F results}
\end{table}

We also evaluate different methods on the L3F dataset~\cite{Lamba2021}, where the LF images were captured under extremely low-light conditions. Table~\ref{tab:L3F results} records the quantitative results of different methods on several scenes and Fig.~\ref{fig:L3F_com} presents some of the visual results, which reflects that our method obtains higher-quality LF images in terms of color, luminance and high-frequency details, and therefore demonstrates stronger generalization compared to the other methods.

\begin{figure}[!hbt]
\centering
\includegraphics[width=0.48\textwidth]{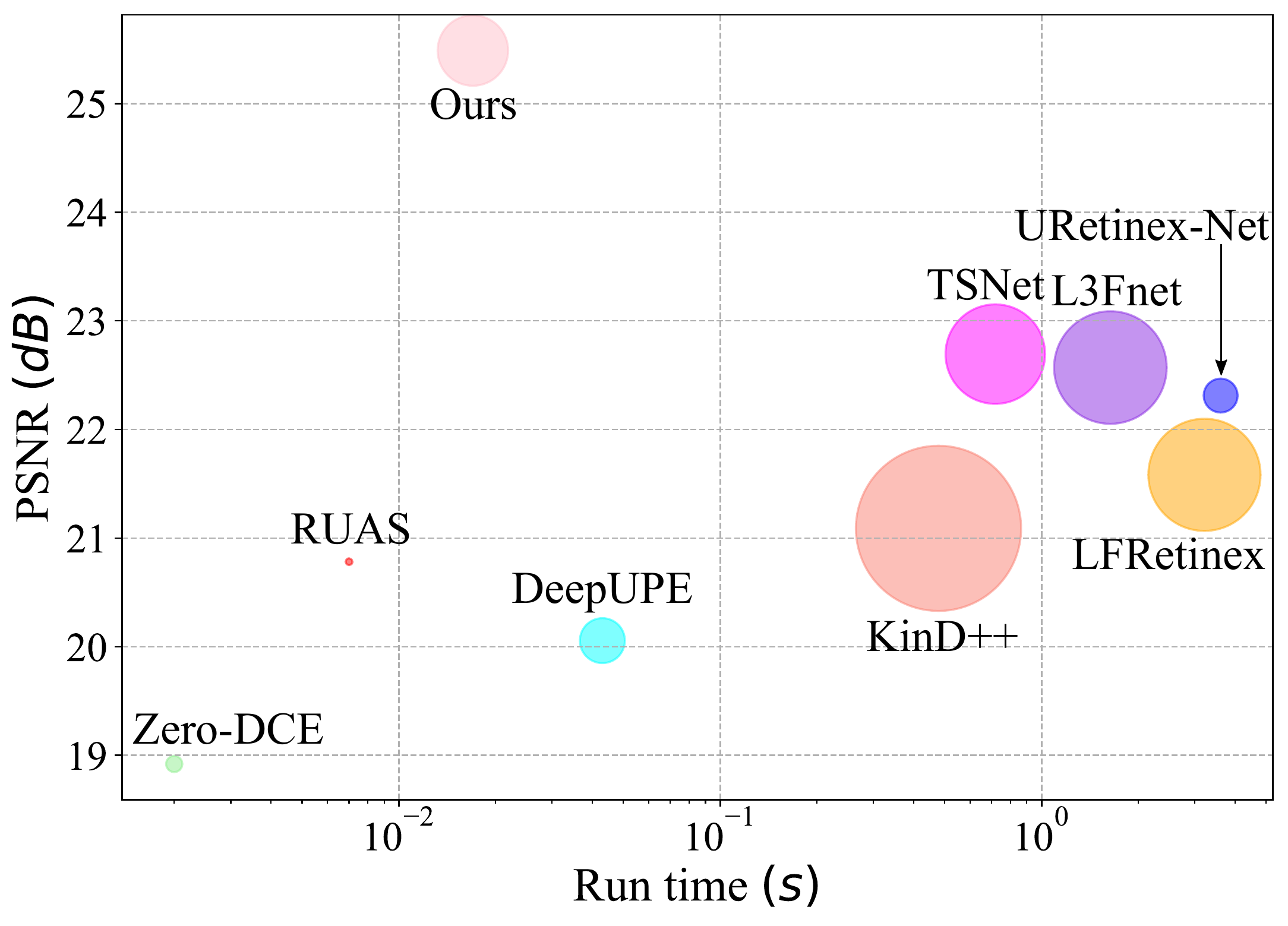} 
\caption{The model performance (PSNR) versus run time of different methods. The circle size of each method is proportional to its model size.}
\label{fig:efficiency}
\end{figure}

\subsubsection{Efficiency}
To evaluate the model efficiency, the average run time and model size of different methods are listed in Table~\ref{tab:Run time}. The run time is measured by restoring a $7\times 7 \times 384\times 512$ dark LF image on P100 GPU. We can see that our method is much more efficient and lightweight than the other LF image methods. Even if Zero-DCE and RUAS have very fast inference speed and small model size, their performances on dark LF restoration are worse than ours. Fig.~\ref{fig:efficiency} presents the model performance in terms of PSNR versus the run time of different methods, which suggests that our method achieves a better balance between performance and efficiency. 

\begin{table}[t]
\caption{Run time and model size of different methods.}
\centering
\begin{tabular}{c|cc}
\toprule
Method  & Run time (s) & Param. (M)\\[0.5ex]
\midrule
DeepUPE~\cite{Wang2019b}     & 0.043 & 0.594 \\
Zero-DCE~\cite{Guo2020}      & 0.002 & 0.079 \\
RUAS~\cite{Liu2021a}         & 0.007 & 0.004 \\
KinD++~\cite{Zhang2021c}     & 0.478 & 8.017 \\
URetinex-Net~\cite{Wu2022}   & 3.609 & 0.340 \\
\midrule
LFRetinex~\cite{Zhang2021b}  & 3.214 & 3.697 \\
L3Fnet~\cite{Lamba2021}      & 1.638 & 3.725 \\
TSNet~\cite{Lamba2022}       & 0.718 & 2.909\\
Ours                         & 0.017 & 1.466 \\
\bottomrule
\end{tabular}
\label{tab:Run time}
\end{table}

\subsubsection{Noise synthesis}

\begin{table}[t]
\caption{Quantitative results on our captured LF images using different noise synthesis methods. Bold: Best.}
\centering
\begin{tabular}{c|ccc}
\toprule
Method & PSNR$\uparrow$ & SSIM$\uparrow$ & LPIPS$\downarrow$\\
\midrule
Random noise~\cite{Zhang2021b} & 25.13 & 0.836 & 0.103 \\
Shot+read noise~\cite{Wang2020d}  & 25.39 & 0.845 & 0.096 \\
Physics-based~\cite{Wei2021} & 23.14 & 0.822 & 0.137\\
Ours & \textbf{26.67} & \textbf{0.857} & \textbf{0.086} \\
\bottomrule
\end{tabular}
\label{tab:noise comparison}
\end{table}

\begin{figure}[t]
\centering
\includegraphics[width=0.48\textwidth]{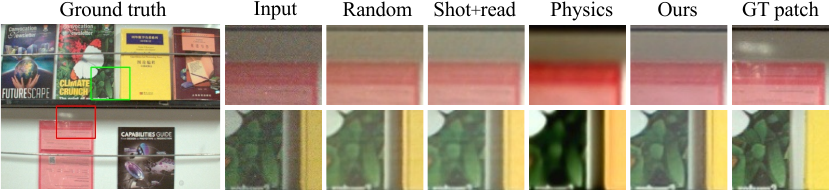} 
\caption{Visual comparison of different noise synthesis methods. The input patches are brightened to clearly show the noise. Zoom in for best view.}
\label{fig:noise_com}
\end{figure}

To validate our noise synthesis method, we made comparison with several other methods, including random Gaussian and Poisson noise used in~\cite{Zhang2021b}, estimated shot noise and read noise as~\cite{Wang2020d}, and a more comprehensive physics-based noise model~\cite{Wei2021}. More specifically, the method in~\cite{Wang2020d} only estimates the shot noise and read noise, and the method in~\cite{Wei2021} absorbs the dark current into the read noise, which is modeled with a Tukey lambda distribution.

We trained our LRT using the LF images synthesized with these different noise synthesis methods and then evaluated on our captured LF images. The quantitative results are recorded in Table~\ref{tab:noise comparison}, which suggests that the network trained by our noise synthesis method achieves better performance than the others. Fig.~\ref{fig:noise_com} presents the visual comparison, where we observe that the results of~\cite{Zhang2021b} and~\cite{Wang2020d} have some residual noise since they only consider the shot noise and read noise without providing a comprehensive modeling for the imaging noise, and the results of~\cite{Wei2021} suffer from color distortion and blurriness as the Tukey lambda distribution with a estimated shape parameter for read noise may not agree with the real noise distribution of LF imaging. Therefore, our noise synthesis pipeline is more effective to simulate the real noise.

\begin{figure}[t]
\centering
\includegraphics[width=0.48\textwidth]{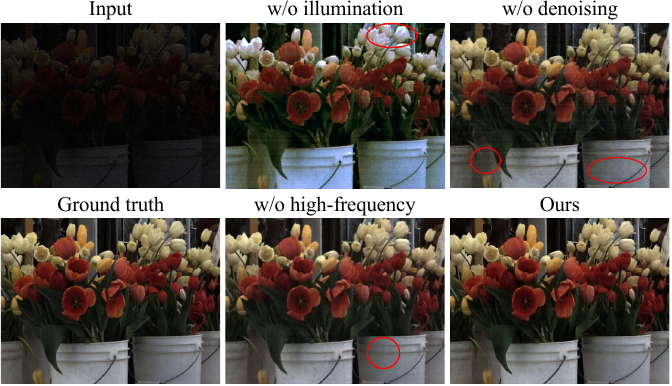} 
\caption{Visual comparison of different configurations. The circled regions reflect the poor performance of the other configurations compared to ours.}
\label{fig:structure_ab}
\end{figure}

\begin{figure}[t]
\centering
\includegraphics[width=0.48\textwidth]{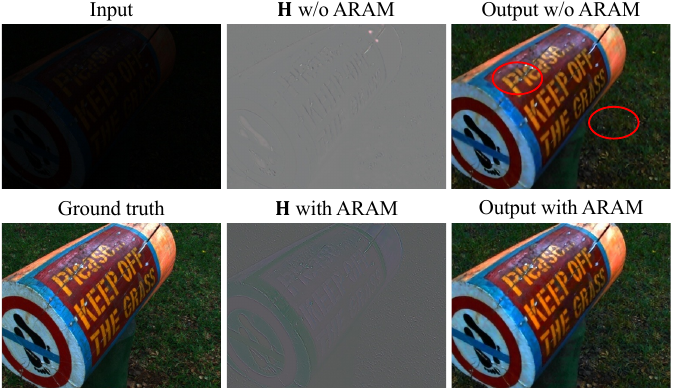} 
\caption{The high-frequency maps and restoration results without and with the adaptive ratio adjustment module (ARAM). The circled regions indicates the blurry details compared to our default output.}
\label{fig:hf_ab}
\end{figure}

\begin{figure}[t]
\centering
\includegraphics[width=0.48\textwidth]{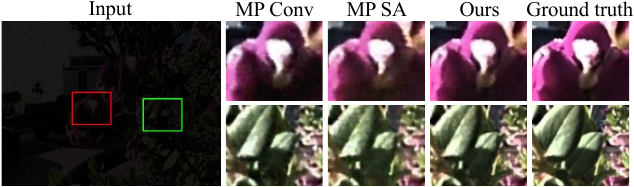} 
\caption{Visual comparison of different angular blocks. Zoom in for best view.}
\label{fig:angular_ab}
\end{figure}

\begin{figure}[t]
\centering
\includegraphics[width=0.48\textwidth]{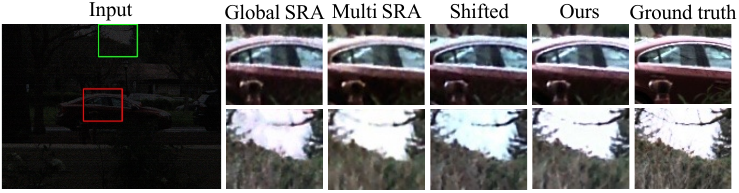} 
\caption{Visual comparison of different spatial blocks. Zoom in for best view.}
\label{fig:spatial_ab}
\end{figure}

\subsection{Ablation Studies}

\begin{table*}[!hbt]
\caption{Ablation study on the network architecture. Bold: Best.}
\centering
\begin{tabular}{cccc|cc|ccc}
\toprule
\multicolumn{4}{c|}{Different heads} & \multicolumn{2}{c|}{Transformer blocks} & \multicolumn{3}{c}{Results} \\
\midrule
Illumination & Denoising & High-frequency & Adaptive ratio adjustment & Angular & Spatial & PSNR$\uparrow$ & SSIM$\uparrow$ & LPIPS$\downarrow$\\[0.5ex]
\midrule
            &            &            &            & \checkmark & \checkmark & 22.74 & 0.741 & 0.199 \\
            &            & \checkmark & \checkmark & \checkmark & \checkmark & 23.16 & 0.747 & 0.181 \\
 \checkmark &            & \checkmark & \checkmark & \checkmark & \checkmark & 24.48 & 0.768 & 0.162 \\
 \checkmark & \checkmark &            &            & \checkmark & \checkmark & 26.31 & 0.824 & 0.124 \\
 \checkmark & \checkmark & \checkmark &            & \checkmark & \checkmark & 26.77 & 0.831 & 0.116 \\
\midrule
 \checkmark & \checkmark & \checkmark & \checkmark &            & \checkmark & 24.62 & 0.764 & 0.151 \\
 \checkmark & \checkmark & \checkmark & \checkmark & \checkmark &            & 25.98 & 0.801 & 0.137 \\
\midrule
 \checkmark & \checkmark & \checkmark & \checkmark & \checkmark & \checkmark & \textbf{27.68} & \textbf{0.842} & \textbf{0.103} \\
\bottomrule
\end{tabular}
\label{tab:architecture ablation}
\end{table*}

\subsubsection{Network architecture}
We conducted several ablation studies to validate our architecture design. The quantitative results of different network configurations on the synthetic \texttt{test} set are recorded in Table~\ref{tab:architecture ablation}. We first trained a model without any head, which is equivalent to the full-scale image-to-image mapping. The results are much worse than our full configuration, verifying the effectiveness of intermediate-task learning. Moreover, we trained a model without the illumination head and denoising head (become direct mapping from $\mathbf{L}_\mathrm{in}^{1/4}$ to $\mathbf{L}_\mathrm{re}^{1/4}$), a model without denoising head, a model without high-frequency head, and a model with high-frequency head but without the adaptive ratio adjustment. We can see that the performance is degraded significantly when learning a direct mapping from $\mathbf{L}_\mathrm{in}^{1/4}$ to $\mathbf{L}_\mathrm{re}^{1/4}$ or not using the denoising head. The performance suffers some decline if the high-frequency head is removed, and also has little decline if the adaptive ratio adjustment module for detail prediction is not incorporated. Then, we validate our angular transformer blocks by removing them and the spatial transformer blocks by replacing them with the residual blocks. The degraded performances verify the validity of modeling the global angular relationship and multi-scale spatial dependencies.

The visual comparison of different configurations is presented in Fig.~\ref{fig:structure_ab}, where we observe that the model without the illumination head and denoising head leads to obvious color distortion, the model without denoising head fails to suppress severe noise, and the model without high-frequency head cannot preserve clear local details compared to our full configuration. In addition, Fig.~\ref{fig:hf_ab} shows the predicted high-frequency maps and their corresponding restoration results with and without the adaptive ratio adjustment module. It can be seen that more and clearer local details can be predicted by introducing adaptive ratio adjustment for the extremely low-light input, therefore obtaining higher-quality output.

\subsubsection{Angular and spatial transformer blocks}

\begin{table}[t]
\caption{Ablation study on the transformer blocks. Bold: Best.}
\centering
\scalebox{0.85}{
\begin{tabular}{c|c|ccc}
\toprule
Configuration & Run time (s) & PSNR$\uparrow$ & SSIM$\uparrow$ & LPIPS$\downarrow$\\
\midrule
MP $3\times 3$ conv & 0.027 & 25.98 & 0.817 & 0.127 \\
MP self-attention & 0.023 & 26.45 & 0.828 & 0.117 \\
\midrule
Global SRA & 0.019 & 26.19 & 0.815 & 0.122 \\
Multi-scale SRA & 0.023 & 26.69 & 0.825 & 0.109 \\
Shifted window & 0.017 & 26.07 & 0.807 & 0.132 \\
\midrule
 Ours & 0.017 & \textbf{27.68} & \textbf{0.842} & \textbf{0.103} \\
\bottomrule
\end{tabular}}
\label{tab:transformer ablation}
\end{table}

We conducted additional ablation studies to validate our angular and spatial transformer blocks. We experimented with another two methods for angular feature extraction, applying $3\times3$ convolution on the macro-pixel (`MP $3\times 3$ conv') used in~\cite{Yeung2019} and computing global self-attention on the macro-pixel (`MP self-attention') used in~\cite{Liang2022}. We replaced our angular transformer blocks with these two kinds of blocks and retrained the network with the same dataset. The run time and quantitative results are listed in Table~\ref{tab:transformer ablation}. It can be seen that our angular transformer block obtains better results with higher inference efficiency than these two blocks. Fig.~\ref{fig:angular_ab} shows the visual comparison of different angular blocks, which reflects that our angular transformer block contributes to better restoration.

For the spatial blocks, we experimented with another three transformer blocks, the global spatial-reduction self-attention (SRA)~\cite{Wang2022a}, the multi-scale SRA~\cite{Ren2022}, and the shifted window self-attention~\cite{Liu2021b}. Both the global SRA and multi-scale SRA incorporate only the global self-attention, while the shifted window incorporates only the window-based local self-attention. We replaced our spatial transformer blocks with these three kinds of blocks to train additional models. The quantitative results in Table~\ref{tab:transformer ablation} and visual comparison in Fig.~\ref{fig:spatial_ab} suggests that our spatial transformer block with multi-scale self-attention is more effective for low-light restoration.

\begin{figure}[t]
\centering
\includegraphics[width=0.48\textwidth]{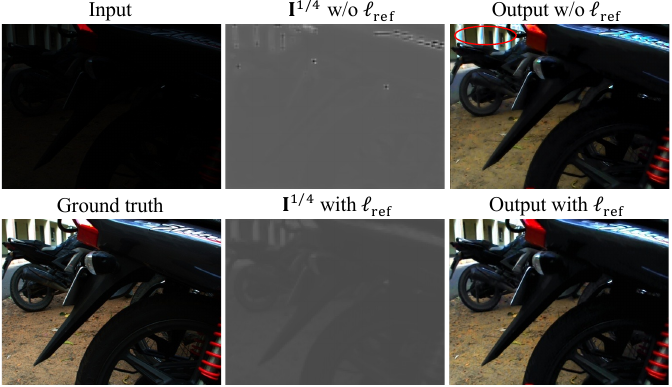} 
\caption{The illumination maps and the restoration results without and with the illumination reference loss.}
\label{fig:loss_ab}
\end{figure}

\subsubsection{Loss terms}

\begin{table}[t]
\caption{Ablation study on the loss terms. Bold: Best.}
\centering
\begin{tabular}{cc|ccc}
\toprule
\multicolumn{2}{c|}{Loss terms} & \multicolumn{3}{c}{Results} \\
\midrule
$\ell_\mathrm{ref}$ & $\ell_\mathrm{SSIM}$ & PSNR$\uparrow$ & SSIM$\uparrow$ & LPIPS$\downarrow$\\
\midrule
           & \checkmark & 26.82 & 0.832 & 0.113 \\
\checkmark &            & 27.14 & 0.836 & 0.105 \\
\midrule
\checkmark & \checkmark & \textbf{27.68} & \textbf{0.842} & \textbf{0.103} \\
\bottomrule
\end{tabular}
\label{tab:loss ablation}
\end{table}

We also validate our loss design by training additional models without the illumination reference loss and SSIM loss, respectively, since the other loss terms are indispensable to achieve the intermediate and main tasks. Table~\ref{tab:loss ablation} lists their quantitative results, which suggests that these two loss terms can help to improve the performance. Fig.~\ref{fig:loss_ab} shows the estimated illumination maps and their corresponding restoration results with and without the illumination reference loss. It can be seen that the illumination map using the reference loss preserves clearer object boundaries and achieves better estimation for the light distribution, which leads to the high-quality output, while the restored image without the reference loss has some unexpected defects due to its poor illumination map, as shown in the circled region.

\section{Conclusion}
In this paper, we propose the LRT, an efficient low-light restoration transformer for LF images, which contains multiple heads to implement denoising, luminance adjustment, refinement and detail enhancement, respectively, achieving progressive restoration from small scale to full scale. In addition, we design an angular transformer block with a view-token scheme to model the global angular dependencies across all the views, and a multi-scale window-based transformer block to extract global and multi-scale local spatial features within each view. In order to synthesize more realistic dark LF images, we estimate the noise parameters of the LF camera under different ISOs and use them to simulate the corresponding noise. Our network was trained on the synthetic dataset and can generalize well to the real low-light scenarios. It outperforms the other state-of-the-art low-light enhancement methods with better quantitative and qualitative results.

\section*{Acknowledgment}
The work is supported in part by the Research Grants Council of Hong Kong (GRF 17201620, 17200321) and by ACCESS --- AI Chip Center for Emerging Smart Systems, Hong Kong SAR.

\bibliographystyle{IEEEtran}
\bibliography{references}

\end{document}